\label{key}

\documentclass[times, review, 10pt]{elsarticle}

\usepackage{setspace}  
\newtheorem{definition}{Definition}
\usepackage[numbers]{natbib}
\usepackage{setspace}  
\usepackage{amsmath}
\usepackage{amssymb}
\usepackage{amsfonts}
\usepackage{booktabs} 
\usepackage{tikz}
\usepackage{chngcntr}  
\usepackage{graphicx}%
\usepackage{caption}
\usepackage{comment}%
\usepackage{array}  
\usepackage[ruled]{algorithm2e}
\usepackage{diagbox} %
\usepackage{arydshln} %
\usepackage{array} %
\usepackage{multirow} %
\usepackage{booktabs} 
\usepackage{url}
\usepackage{bm}  %
\usepackage{lmodern}  %
\usepackage{float}
 
\usepackage{amssymb}

\journal{arXiv}

\begin{document}

\begin{frontmatter}
	\boldmath   
	
	
	\title{Orthogonal Factor-Based Biclustering Algorithm (BCBOF) for High-Dimensional Data and Its Application in Stock Trend Prediction}
	
	
	\author{Yan Huang}
	\author{Da-Qing Zhang\corref{cor1}}
	\ead{d.q.zhang@ustl.edu.cn}
	\cortext[cor1]{Corresponding author.}
	
	\address{School of Science, University of Science and Technology Liaoning, Anshan, Liaoning, 114051, PR China}

	\begin{abstract}
		Biclustering is an effective technique in data mining and pattern recognition. Biclustering algorithms based on traditional clustering face two fundamental limitations when processing high-dimensional data: (1) The distance concentration phenomenon in high-dimensional spaces leads to data sparsity, rendering similarity measures ineffective;
		(2) Mainstream linear dimensionality reduction methods disrupt critical local structural patterns. To apply biclustering to high-dimensional datasets, we propose an orthogonal factor-based biclustering algorithm (BCBOF).
		First, we constructed orthogonal factors in the vector space of the high-dimensional dataset. Then, we performed clustering using the coordinates of the original data in the orthogonal subspace as clustering targets. Finally, we obtained biclustering results of the original dataset. Since dimensionality reduction was applied before clustering, the proposed algorithm effectively mitigated the data sparsity problem caused by high dimensionality.
		Additionally, we applied this biclustering algorithm to stock technical indicator combinations and stock price trend prediction. Biclustering results were transformed into fuzzy rules, and we incorporated profit-preserving and stop-loss rules into the rule set, ultimately forming a fuzzy inference system for stock price trend predictions and trading signals.
		To evaluate the performance of BCBOF, we compared it with existing biclustering methods using multiple evaluation metrics. The results showed that our algorithm outperformed other biclustering techniques.
		To validate the effectiveness of the fuzzy inference system, we conducted virtual trading experiments using historical data from 10 A-share stocks. The experimental results showed that the generated trading strategies yielded higher returns for investors.
		
	\end{abstract}
	
	\begin{keyword}
		Biclustering\sep Stock volatility prediction\sep High-dimensional data\sep Fuzzy inference system\sep Technical indicators\sep Stock trading strategies
		
		
	\end{keyword}
	
\end{frontmatter}


\section{Introduction}\label{sec1}
 
Since its initial proposal by Cheng and Church\cite{cheng_biclustering_2000}, biclustering has evolved into a sophisticated analytical approach. Conventional clustering methods, while effective for global pattern recognition, exhibit inherent limitations in capturing localized data structures. Crucially, these local patterns often contain biologically significant information that may be obscured in global analyses. Biclustering techniques were specifically developed to address this limitation by simultaneously identifying localized patterns within defined subsets of genes and experimental conditions.

Biclustering algorithms identify local data patterns - an NP-hard problem. The foundational CC algorithm \cite{cheng_biclustering_2000} detects correlated gene-condition clusters through iterative matrix updates, though its computational efficiency is constrained by row/column evaluations, especially for large/incomplete datasets. To address this, the FLexible Overlapped biclustering (FLOC) \cite{yang_enhanced_2003} incorporates probabilistic missing-value modeling to enable overlapping bicluster detection. For large-scale gene expression data, the Iterative Signature Algorithm (ISA) \cite{bergmann_iterative_2002} and the Quantitative Biclustering (QUBIC) \cite{li_qubic_2009} achieve linear-time iterations relative to gene/condition counts, ensuring scalable performance. The recently developed Rapid Unsupervised Biclustering (RUBic) algorithm \cite{sriwastava_rubic_2023} demonstrates significant improvements in processing efficiency for massive datasets compared to existing methods.

When analyzing complex or noisy datasets, traditional clustering methods frequently fail to reveal underlying patterns effectively. Researchers have consequently developed advanced algorithms to enhance clustering performance \cite{lazzeroni_plaid_2002, gu_bayesian_2008,  prelic_systematic_2006, kluger_spectral_2003, izenman_local_2011}.
The plaid model-based algorithm proposed by Lazzeroni \cite{lazzeroni_plaid_2002} decomposes datasets into superimposed clustering patterns, particularly effective for identifying overlapping biclusters, though its greedy nature may constrain practical utility. Gu's Bayesian adaptation \cite{gu_bayesian_2008} eliminates threshold requirements while maintaining detection accuracy through probabilistic modeling.
Alternatively, Prelic's Bimax algorithm \cite{prelic_systematic_2006} employs divide-and-conquer optimization for binary data models. Parameter estimation methods \cite{kluger_spectral_2003, izenman_local_2011} detect patterns via statistical modeling, albeit with computational intensity and parameter sensitivity.
The algorithms mentioned above are all capable of effectively capturing nonlinear relationships in data and enhancing the model's ability to recognize complex patterns.

Beyond computational efficiency, enhancing the interpretability of biclustering results has become a crucial research focus. Denitto significantly advanced this field through two innovative contributions \cite{denitto_spike_2017, denitto_biclustering_2020}: the 2017 sparse low-rank matrix decomposition method that improves feature selection, noise robustness, and interpretability; and the 2020 graph-based reformulation using dominant sets that enables simultaneous row-column analysis.
Biclustering has now evolved into a diverse research field. These approaches include biclustering algorithms based on traditional clustering \cite{qu_constrained_nodate, getz_coupled_2000}, greedy iterative search \cite{cheng_biclustering_2000, yang_enhanced_2003, bergmann_iterative_2002, li_qubic_2009}, exhaustive biclustering strategies \cite{liu_op-cluster_2003}, mathematical model approaches \cite{lazzeroni_plaid_2002, gu_bayesian_2008}, and other methods \cite{denitto_spike_2017, denitto_biclustering_2020, sriwastava_rubic_2023, bryan_biclustering_2005, divina_multi-objective_nodate}.

Biclustering algorithms have emerged as essential tools in financial market research \cite{liu_biclustering_2022, huang_automated_2020}, effectively addressing the inherent volatility and uncertainty that generate investment risks. While classical technical indicators \cite{huang_automated_2020, wang_dynamical_2015, wang_dynamical_2015-2, wang_dynamical_2015-1} remain widely employed, their oversimplified structures frequently result in prediction inaccuracies. Although combining multiple indicators enhances precision, the optimal selection and customization for individual stocks presents ongoing challenges. Biclustering algorithms can more accurately capture the market's nonlinear characteristics and potential trends by considering multiple trading day indicator values and technical indicator combinations simultaneously. Therefore, using biclustering, we can achieve the combination of stock technical indicators and the pattern recognition of fluctuation trends based on this combination.

However, the abundance of technical indicators and the extensive volume of historical trading data present significant challenges. When employing biclustering algorithms to extract local indicator combinations and trend patterns, the method encounters the fundamental issue of high-dimensional data. In such high-dimensional spaces, data points tend to exhibit uniformly distributed distances, resulting in inherent sparsity. This sparsity phenomenon substantially compromises the effectiveness of distance-based clustering algorithms. Specifically, in high-dimensional environments, clustering algorithms face considerable difficulties in reliably identifying the underlying cluster structures, ultimately leading to compromised accuracy and stability in the biclustering results. Biclustering algorithms based on traditional clustering often experience significant impacts on their effectiveness when dealing with high-dimensional data. Biclustering algorithms based on traditional clustering encounter fundamental limitations when processing high-dimensional data: (1) The concentration of distances in high-dimensional space leads to data sparsity, causing distance-based similarity measures to become ineffective; (2) Mainstream linear dimensionality reduction methods, such as PCA, impose global low-dimensional projections that disrupt the local manifold structure of the data, which is a key pattern feature that biclustering aims to detect. Orthogonal factor models, by constructing local orthogonal bases and domain-adapted factor interpretations, can significantly alleviate the local structure loss problem inherent in PCA-like methods, making them particularly suitable for data with clear economic meanings, such as financial technical indicators. Therefore, to improve the accuracy of biclustering in high-dimensional data, we introduce a dimensionality reduction technique based on the orthogonal factor model to construct orthogonal subspace. The constraints of orthogonal factors make the clustering results more interpretable and stable. Compared to other biclustering methods, our approach is better at preserving the intrinsic structure of the data and effectively reducing redundant information when handling high-dimensional data. This method not only simplifies the complexity of data processing but also significantly enhances the effective identification capability of indicator combinations. Based on this, we explore the technical trading patterns underlying the combinations of indicators in depth. These patterns are crucial for gaining insights into market dynamics. We utilize the "IF-THEN" rules within a fuzzy inference system to convert the mined trading patterns into stock trend predictions. To enhance the overall profitability and security of stock trading, we have designed profit-preserving and stop-loss rules that incorporate time delay strategy, buying price control, and loss rate threshold control. By integrating historical trading experience-based rules with stock trend predictions, we have formulated a flexible stock trading strategy. This strategy specifies precise timing for buying or selling under three different market conditions, ensuring the effectiveness of trading decisions.

The main innovations of this paper are two-fold:

(1) We propose an improved biclustering algorithm that reduces the dimensionality of high-dimensional data through an orthogonal factor model. This algorithm reduces the complexity of clustering high-dimensional data and enhances algorithm stability, thereby improving the accuracy of biclustering results.

(2) When applied to stock data, the algorithm identifies trading patterns based on technical indicators. These identified trading patterns are described by  using fuzzy rules and are combined with profit-preserving and stop-loss rules to derive a stock trading strategy. This strategy can provide buy or sell signals for stocks, including timing and price of the trade.

The remainder of the paper is as follows. Section \ref{sec2} introduces the work related to the proposed method. Section \ref{sec3} provides a detailed introduction to the proposed method and its application to the combination of technical indicators. Section \ref{sec4} presents the experimental results of the proposed method and compares it with other methods. Section \ref{sec5} concludes the method proposed.

\section{Related Work}\label{sec2}
\subsection{Biclustering algorithm}
The concept of biclustering was proposed by Cheng and Church \cite{cheng_biclustering_2000}, intended for mining gene expression patterns in gene expression data analysis. The principle allows for the automatic discovery of similarity based on attribute subsets. It simultaneously clusters genes and conditions. 
The purpose of biclustering is to seek submatrices in the gene expression data matrix that satisfy the conditions. The gene set in the submatrix is consistently expressed on the corresponding condition set. Biclustering algorithm is a type of algorithm used to discover submatrices with similar patterns in the data matrix, which could be applied in the fields of gene expression data, proteomics data, and so on.

Let $A$ be an $n \times p$ matrix that represents a dataset, where $a_{ij}$ denotes the element in the $i$-th row and the $j$-th column. The biclustering problem can be defined as finding a submatrix $B$ within the matrix $A$, such that the mean square residual score of submatrix $B$ is less than a given threshold.  

\begin{definition}
	Give $\delta \ge 0$, a submatrix $B$ of the matrix $A$ is called a $\delta$-bicluster, if $H\left ( I,J \right ) \le \delta$. The mean square residual score $H(I, J)$ of the submatrix $B$ is defined as: 
\end{definition}
\begin{equation}
	H(I, J)=\frac{1}{|I||J|} \sum_{i \in I, j \in J}\left(r_{i j}\right)^{2}=\frac{1}{|I||J|} \sum_{i \in I, j \in J}\left(a_{i j}-a_{i J}-a_{Ij}+a_{IJ}\right)^{2}\label{MSR}
\end{equation}
\begin{equation}
	a_{i J}=\frac{1}{|J|} \sum_{j=J} a_{i j}, a_{i j}=\frac{1}{|I|} \sum_{i \in I} a_{i j}
\end{equation}
\begin{equation}
	a_{I J}=\frac{1}{|I||J|} \sum_{i=I,j=J} a_{i j}=\frac{1}{|I|} \sum_{i \in I} a_{i J}=\frac{1}{|J|} \sum_{j \in J} a_{I j}
\end{equation}
where, $I$ is a subset of $\left \{1, 2,\cdots, n  \right \} $, $J$ is a subset of $\left \{1, 2,\cdots, p  \right \} $. $|I|$ denotes the number of elements in $I$, and $|J|$ is the number of elements in $J$. $r_{ij}$ represents the residual of the element $a_{ij}$, where $a_{iJ}$, $a_{Ij}$, and $a_{IJ}$ denote the row mean value, column mean value, and the mean value of the submatrix $B$, respectively.

A smaller residual value suggests a higher correlation of the element with its row and column, and therefore it is more likely to be classified into the same category during clustering. On the contrary, a larger residual value suggests a substantial difference between the element and the other elements. Therefore, residuals can serve as a key measure of the correlation between an element and the other elements in a bicluster. The mean square residual $H(I, J)$ is used to evaluate the quality and precision of bicluster. It reflects the degree of deviation between the data points within the submatrix and the average of the submatrix. A smaller mean square residual implies that the data points within the submatrix are closer to their average, thereby reflecting better clustering effects and similarity between data points.

From the above definition, it is clear that due to the mean square residual of bicluster is less than a threshold and does not require optimization, there may exist multiple biclusters satisfying these conditions. These biclusters may exhibit different patterns,  known as multimodal. This implies that there are multiple patterns in the dataset, and each pattern corresponds to a bicluster. The results of $\delta$-bicluster are typically not unique, allowing us to obtain multiple fuzzy rules.

\section{Proposed Approach}\label{sec3}
\subsection{Biclustering based on orthogonal factor} 
Biclustering algorithms based on traditional clustering are often difficult to obtain accurate clustering results when dealing with large data matrices due to the excessive number of rows and columns. When the matrix is high-dimensional, meaning the data to be biclustered is high-dimensional, the mean square residual score $H(I,J)$ becomes less sensitive to the threshold $\delta$ as the dimensionality of the data and the number of samples increase, leading to inaccurate biclustering results. When performing biclustering on high-dimensional data, it is essential to apply appropriate dimensionality reduction to the data. The orthogonal factor model is an effective method for data dimensionality reduction.

The orthogonal factor model appears in the factor analysis \cite{galbraith_analysis_nodate} process. Factor analysis is a multivariate statistical method that describes the relationships among multiple variables by constructing a few factors, thereby achieving dimensionality reduction and data simplification. Its basic idea is to categorize variables. Those variables with high correlation and closeness are classified into the same category. The variables among different categories have low correlation, and therefore each category of variables represents a common factor. Factor analysis can also be used for classification processing of variables or samples. According to the factor loading matrix, variables or sample points are represented in the space composed of factor axes, achieving the purpose of classification.

Let $X$ be an $n \times p$ matrix,
\begin{equation}
	X=\left[\begin{array}{cccc}
		x_{11} & x_{12} & \cdots & x_{1 p} \\
		x_{21} & x_{22} & \cdots & x_{2 p} \\
		\vdots & \vdots & \vdots & \vdots \\
		x_{n 1} & x_{n 2} & \cdots & x_{n p}\end{array}\right]=\left(\begin{array}{c}
		x_{1} ,x_{2} ,\dots ,x_{p}
	\end{array}\right)
	\label{4} 
\end{equation}
where, $x_{i}$ denotes the $i$-th column of matrix $X$.

When performing factor analysis on the columns of this matrix, the orthogonal factor model is as follows:
\begin{equation}
	\left\{\begin{array}{c}x_{1}=a_{11} F_{1}+a_{12} F_{2}+\cdots+a_{1 m} F_{m}+\varepsilon_{1} \\x_{2}=a_{21} F_{1}+a_{22} F_{2}+\cdots+a_{2 m} F_{m}+\varepsilon_{2} \\\vdots \\x_{p}=a_{p 1} F_{1}+a_{p 2} F_{2}+\cdots+a_{p m} F_{m}+\varepsilon_{p}\end{array}\right.\label{eq:einstein} 
\end{equation}
Since model (\ref{eq:einstein}) is designed for variables and the factors are orthogonal, it is also known as the $R$-type orthogonal factor model. In matrix form, it is expressed as: $X=A F+\varepsilon$. $F=(F_{1},\cdots,F_{m})$ is referred to as the common factors of $X$, $\varepsilon=(\varepsilon_{1},\cdots,\varepsilon_{p})$ is called the specific factors of $X$, the matrix $A=(a_{ij})\in R^{p\times m}$ is known as the factor loading matrix, and $a_{ij}$ is called the factor loading, representing the correlation coefficient between the $i$-th variable and the $j$-th factor.

In the orthogonal factor model (\ref{eq:einstein}), the common factors $F_{i},i=1, 2, \dots, m$, are orthogonal to each other, and they form an orthogonal basis for an inner product subspace within the space occupied by $x_{i},i=1, 2, \dots, p$ in (\ref{4}). The factor loadings $(a_{i,1},a_{i,2},\dots,a_{i,m})$ can be interpreted as the coordinates of $x_i$ in the orthogonal basis $(F_1,\dots,F_m)$. When $(F_1,\dots,F_m)$ are given, there is a one-to-one correspondence between $x_i$ and $(a_{i,1},\dots,a_{i,m})$ for all $i=1,\dots,p$. This allows us to transform the original clustering problem of $p$ $n$-dimensional vectors into the clustering of $p$ $m$-dimensional vectors.

Let's assume that $p$ represents the number of technical indicators (e.g., 30 technical indicators), and $n$ represents the number of trading days for a particular stock (e.g., 200 trading days). When clustering technical indicators, we would be faced with the task of clustering 30 vectors of 200 dimensions. Clustering algorithms inevitably involve the use of distances between vectors. However, in high-dimensional spaces, vectors tend to become sparse, making it difficult to determine whether points are ``close'' to each other based on ``distance''.

The orthogonal factor model provides a solution to this problem by enabling the clustering of $n$-dimensional vectors through $m$-dimensional vectors, namely the factor loadings. The steps of finding factor loadings are as follows:

Construct factor variables. Based on the sampled data, we use the principal component method to construct factor variables and estimate the factor loading matrix $A$.

Factor rotation. Due to the difficulty in explaining the practical significance of the initially calculated common factors, it is necessary to rotate the factor loading matrix. The commonly used method is the maximum variance (Varimax) \cite{kaiser_varimax_1958} method. We obtain the rotated factor loading matrix $B$ by applying Varimax rotation to the factor loading matrix $A$.

Applying agglomerative hierarchical clustering \cite{Ran2023} to the rotated factor loading matrix $B$, we are essentially calculating the distances between each column $x_i,i=1, \cdots, p$, on the data matrix $X$ based on their coordinate positions in the new orthogonal coordinate system. This approach reveals the similarities and differences between each column, and the clustering results are obtained accordingly. Agglomerative hierarchical clustering is a bottom-up clustering method that initially treats each column as a separate cluster and then gradually merges similar clusters until a certain stopping criterion is met. By applying agglomerative hierarchical clustering to the factor loading matrix, we can obtain clustering results for the columns of the data matrix. Since the factor loading matrix has already experienced  dimensionality reduction, agglomerative hierarchical clustering is able to more accurately identify the similarities and differences between columns, resulting in more precise clustering outcomes.

After the column clustering based on orthogonal factors, we obtain multiple submatrices that contain all rows and some columns of the original data matrix. Next, row clustering is performed on these submatrices to form a biclustering structure, which considers the similarity of both rows and columns. The purpose of biclustering is to identify regions in the data matrix where both rows and columns exhibit strong correlations.

We employ DBSCAN \cite{ester_density-based_nodate} for row clustering, which features automatic cluster detection, strong noise robustness, adaptability to complex distributions, and subspace clustering capabilities. These features make it particularly effective for identifying strong row-column correlations in noisy datasets with complex local patterns, enhancing bicluster interpretability. Comparative evaluations against other algorithms (Tables \ref{Comparison of clustering algorithms} and \ref{DBSCAN}) demonstrate DBSCAN's superior performance on metrics including Silhouette Score, Fowlkes-Mallows Index, and Completeness.
\begin{table}[htbp]
	\centering
	\scriptsize
	\caption{Comparative Characteristics of Clustering Algorithms}
	\setlength{\tabcolsep}{3pt}
	\begin{tabular}{lccccc}
		\hline
		\toprule
		Method & Requires K & Noise Robustness & Arbitrary Shape & Efficiency & High-Dim  \\ 
		\midrule
		K-means\cite{macqueen_methods_nodate} & Yes & Weak & No & High & Fair   \\
		Hierarchical\cite{murtagh_algorithms_2012, johnson_hierarchical_1967} & Indirect & Weak & Yes & Low & Fair  \\
		GMM\cite{reynolds2009gmm} & Yes & Medium & No & Medium & Fair  \\
		DBSCAN\cite{ester_density-based_nodate, hahsler_dbscan_2019, hutchison_densitybased_2013} & \textbf{No} & \textbf{Strong} & \textbf{Yes} & \textbf{High} & \textbf{Good}  \\ 
		\bottomrule
		
	\end{tabular}
	\label{Comparison of clustering algorithms}
	\begin{flushleft}
		\footnotesize{$^*$ ``K'' represents the number of clusters.}
	\end{flushleft}
\end{table}
\begin{table}[htbp]
	\centering
	
	\caption{Comparison results of DBSCAN on three metrics}
	\scriptsize
	\begin{tabular}{lccc}
		\hline
		\toprule
		Method & Silhouette Score & Fowlkes-Mallows Index & Completeness Score  \\ 
		\midrule
		K-means & 0.1868 & 0.3171 & \bf{0.2691}   \\
		Hierarchical & 0.1423 & 0.2983 & 0.2451  \\
		GMM& -0.1423 & 0.2689 & 0.1421    \\
		DBSCAN & \bf{0.5370} & \bf{0.4666} & 0.2518  \\ 
		\bottomrule
		
	\end{tabular}
	\label{DBSCAN}
\end{table}

From Table \ref{DBSCAN}, it is evident that the DBSCAN algorithm performs excellently across multiple metrics, demonstrating its effectiveness. For each submatrix obtained from the previous step, we employ DBSCAN clustering to obtain multiple biclusters.
The pseudo-code for biclustering based on orthogonal is proposed in Algorithm 1.
\begin{algorithm}[]  
	\renewcommand{\thealgocf}{1}
	\SetAlgoLined  
	\caption{BCBOF (Biclustering based on orthogonal factors)}\label{alg:BCBOF} 
	\LinesNumbered 
	\KwIn{Data matrix $X$, threshold $\delta$, radius $\varepsilon$, $minPts$ \tcc{$\varepsilon$ is the neighborhood radius of DBSCAN, $minPts$ is the minimum number of sample points required for DBSCAN}} 
	\KwOut{$Bicluster$;}%
	
	the factor loading matrix $A$ $\leftarrow$ the principal component method in (\ref{eq:einstein})\;
	
	the rotated factor loading matrix $B$ $\leftarrow$ Varimax($A$)\;
	
	$clusters$ $\leftarrow$ agglomerative hierarchical clustering($B$)\;
	\For{i, cluster in enumerate(clusters)}{
		
		cluster labels $\leftarrow$ DBSCAN($cluster$, $\varepsilon$, $minPts$)\;
		
		unique $labels$ $\leftarrow$ get\_unique\_labels (cluster labels)\;
		unique $labels$.discard(-1) $\leftarrow$ remove the label of the noise point \;
		row $indices$ list $\leftarrow$ empty list ()\;
		\For{each cluster label and its corresponding row indices}{ 
			\If{the label is not a noise point}{
				add the row indices to the list corresponding to the label\;
			} 
		} 
		initialize an empty list to store bicluster information\;
		\For{each list of row indices for each bicluster}{
			\If{ the list is not empty}{
				convert the cluster data into a matrix\;								
				the matrix is a bicluster and put it into the set of $Bicluster$\;
			}
		}
	}
\end{algorithm}

\subsection{Combination of stock technical indicators based on algorithm BCBOF} 
We refer to this A stock trading strategy based on the BCBOF algorithm as the SSOBC algorithm, and its workflow is illustrated in the following figure \ref{table}. Sections \ref{3.2.1} through \ref{3.2.3} present a comprehensive description of the SSOBC algorithm's operational procedure.
\begin{figure}[htbp]
	\centering
	\includegraphics[height = 7cm,width = 12cm]{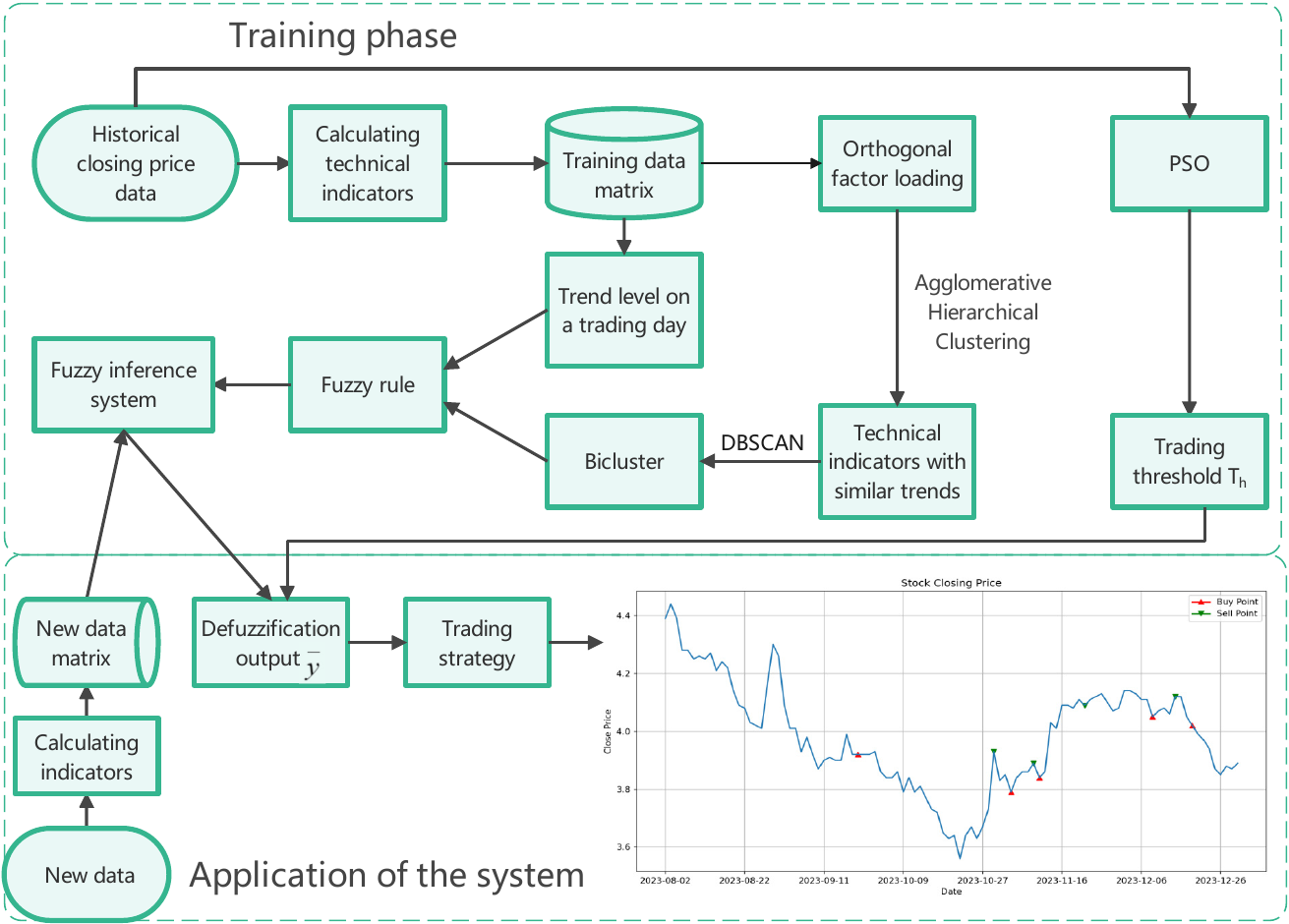}
	\caption{The algorithm proposed in this paper comprises two phases: the training phase and the system application phase. During the training phase, we utilize the BCBOF algorithm in conjunction with a fuzzy inference system to generate fuzzy rules and employ the PSO algorithm to obtain the trading threshold $T_h$. Subsequently, we apply this system to fuzzify new data, ascertain whether it conforms to the established fuzzy rules, and derive a defuzzified value $\bar{y}$. We apply $\bar{y}$ to the proposed trading strategy to generate trading signals.}
	\label{table}
\end{figure}

\subsubsection{Preprocessing of training data}\label{3.2.1}
Before applying the BCBOF algorithm for biclustering on the stock data, we need to preprocess the data. We use historical stock data to form an $n\times p$ data matrix comprised of $n$ trading days, $p$ technical indicators. 
The raw historical stock data may include the opening price, the closing price, the highest price, the lowest price, and the trade volume of a trading day. 
Technical indicators are tools developed through statistical calculations on historical data such as stock price and trade volume, serving to reveal trends in stock price and trading signals. 
There are many technical indicators for stocks, as many as over 100 related technical indicators \cite{Achelis2000}.
We choose $p$ technical indicators from them, such as $ROC$ (Rate of Change), and the time period of which could be selected as 6, 12, 24, 28. 
Because the values of different technical indicators may vary greatly, it is necessary to normalize the data for each column, normalizing the data to the range of [0,1]. 
The data normalization process for each column is as follows:
\begin{equation}
	V'\left ( i,j \right ) =    \frac{V\left ( i,j \right )-V_{min}\left ( j \right )}{V_{max}\left ( j \right )-V_{min}\left ( j \right )}, i=1, \cdots, n , j=1, \cdots, p 
	\label{6}
\end{equation}
where, $V'\left(i,j\right)$ is the value after the technical indicator being normalized, $V\left(i,j\right)$ is the value of the $j$-th technical indicator on the $i$-th day, $V_{max}\left(j\right)$ is the maximum value of the $j$-th technical indicator on $n$ trading days, $V_{min}\left(j\right)$ is the minimum value of the $j$-th technical indicator on $n$ trading days. 

The normalized 244 $\times$ 8 antecedent data matrix is shown in Table \ref{tab:data_matrix}.

\begin{table}[htbp]
	\centering
	\caption{Antecedent Data Matrix}
	\label{tab:data_matrix}
	\footnotesize
	\begin{tabular*}{\textwidth}{@{\extracolsep{\fill}} ccccccccc  @{}}
		\toprule
		Date & $ROC_6$ & $ROC_{12}$ & $ROC_{24}$ & $ROC_{28}$ & $MTM$ & $RSV$  & $AR$ & $BR$  \\
		\midrule
		2022/03/01 & 0.267 & 0.605 & 0.575 & 0.167 & 0.098 & 0.356 & 0.087 & 0.140  \\
		2022/03/02 & 0.284 & 0.623 & 0.663 & 0.211 & 0.121 & 0.304 & 0.107 & 0.183 \\
		2022/03/03 & 0.296 & 0.661 & 0.783 & 0.247 & 0.149 & 0.299 & 0.126 & 0.044  \\
		2022/03/04 & 0.309 & 0.652 & 0.692 & 0.285 & 0.178 & 0.330 & 0.145 & 0.157 \\
		2022/03/07 & 0.315 & 0.688 & 0.659 & 0.335 & 0.202 & 0.469 & 0.171 & 0.141 \\
		2022/03/08 & 0.310 & 0.674 & 0.656 & 0.349 & 0.211 & 0.369 & 0.187 & 0.168  \\
		$\vdots$ & $\vdots$ & $\vdots$ & $\vdots$ & $\vdots$ & $\vdots$ & $\vdots$ & $\vdots$ & $\vdots$ \\
		2023/03/01 & 0.308 & 0.670 & 0.667 & 0.375 & 0.224 & 0.484 & 0.195 & 0.261 \\
		\bottomrule
	\end{tabular*}
\end{table}

In order to capture the trend of stock prices in the historical data, we calculate the change rate of the stock price $CR_i$ for each future trading day. 
We need the average closing price $ACP_i$ for the $n$ trading days after the $i$-th trading day. Let $ACP_{i}$ be defined as follows:
\begin{equation}
	ACP_{i}=\frac{1}{n}  {\textstyle \sum_{m=i}^{i+n-1}} CP_{m}
	\label{7}
\end{equation}
where, $CP_i$ is the closing price on the $i$-th trading day. 
Subsequently, the change rate $CR_i$ of the stock price for each trading day can be calculated as:
\begin{equation}
	CR_{i}=\frac{ACP_{i}-CP_{i}}{CP_{i}}\times 100\%  
	\label{8} 
\end{equation}
where, $CR_i$ is the change rate of the closing price over $n$ trading days after the $i$-th trading day. If $CR_i >$ 0, it implies that the stock price will rise in the future $n$ trading days. If $CR_i < $ 0, it implies that the stock price will fall in the future $n$ trading days.

\begin{table*}
	\centering
	\caption{The $i$-th Trading Day Trend Level}
	\footnotesize
	\begin{tabular*}{\textwidth}{@{\extracolsep{\fill}} ccc  @{}}
		\toprule
		\textbf{Trading day trend level} & \multicolumn{2}{c}{\textbf{Trend$_i$}}   \\
		\midrule
		& if $CR_{i} \geq 3T$ &then $Trend_{i}$ = 3 \\
		Upward &if $2T \leq CR_{i} < 3T$  &then $Trend_{i}$ = 2 \\
		& if $T \leq CR_{i} < 2T$ & then $Trend_{i}$ = 1 \\
		\cdashline{1-3} 
		\addlinespace[0.5em]  
		Stable &if $-T \leq CR_{i}< T$ &then $Trend_{i}$ = 0 \\
		\cdashline{1-3} 
		\addlinespace[0.5em] 
		&if $-2T \leq CR_{i} < -T$ &then  $Trend_{i}$ = -1 \\
		Downward &if $-3T \leq CR_{i}< -2T$ &then  $Trend_{i}$ = -2 \\
		& if $CR_{i} < -3T$ &then $Trend_{i}$ = -3 \\
		\bottomrule
	\end{tabular*}
	\label{Trend Table}
\end{table*}

Regarding the determination of the trend level of the stock price change, there are usually several methods of trend delineation in the reference \cite{huang_automated_2020,zhang_novel_2018}. We use the methodology in reference \cite{huang_automated_2020},
defining a threshold $T$ to judge the trend level of stock price changes. If $T$ is set to 0.5\%, the obtained trend level will serve as the stock trend represented by technical indicators. The trend of trading days is divided into three categories: upward trend, stable trend, and downward trend. The trend is determined by comparing the stock price change rate $CR_i$ with the threshold $T$. Table \ref{Trend Table} serves as the trend level table for the $i$-th trading day, denoted by Trend$_{i}$.

The trend classification based on daily change rate is defined as follows: an upward trend is identified when the change rate exceeds threshold $T$ ($Trend_{i}$ is set between 1 and 3, where higher values indicate stronger upward momentum); a downward trend is determined when the change rate falls below -$T$ ($Trend_{i}$ is set between -1 and -3, with values reflecting the magnitude of decline); and a stable trend is recognized when the absolute change rate is within $T$ ( $Trend_{i}$=0). These trend values are subsequently employed to construct the consequent of fuzzy rules.

\subsubsection{Combination of indicators and pattern recognition}\label{3.2.2}
For the normalized $n\times p$ antecedent data matrix obtained from (\ref{6}), we utilize the BCBOF algorithm to obtain biclusters. Based on the biclustering results, we can discover specific technical trading patterns. Each pattern comprises a particular combination of technical indicators and a corresponding subset of trading day trend values ($Trend_{i}$).
\begin{figure}[htbp]
	\centering
	\includegraphics[height = 6cm,width = 12cm]{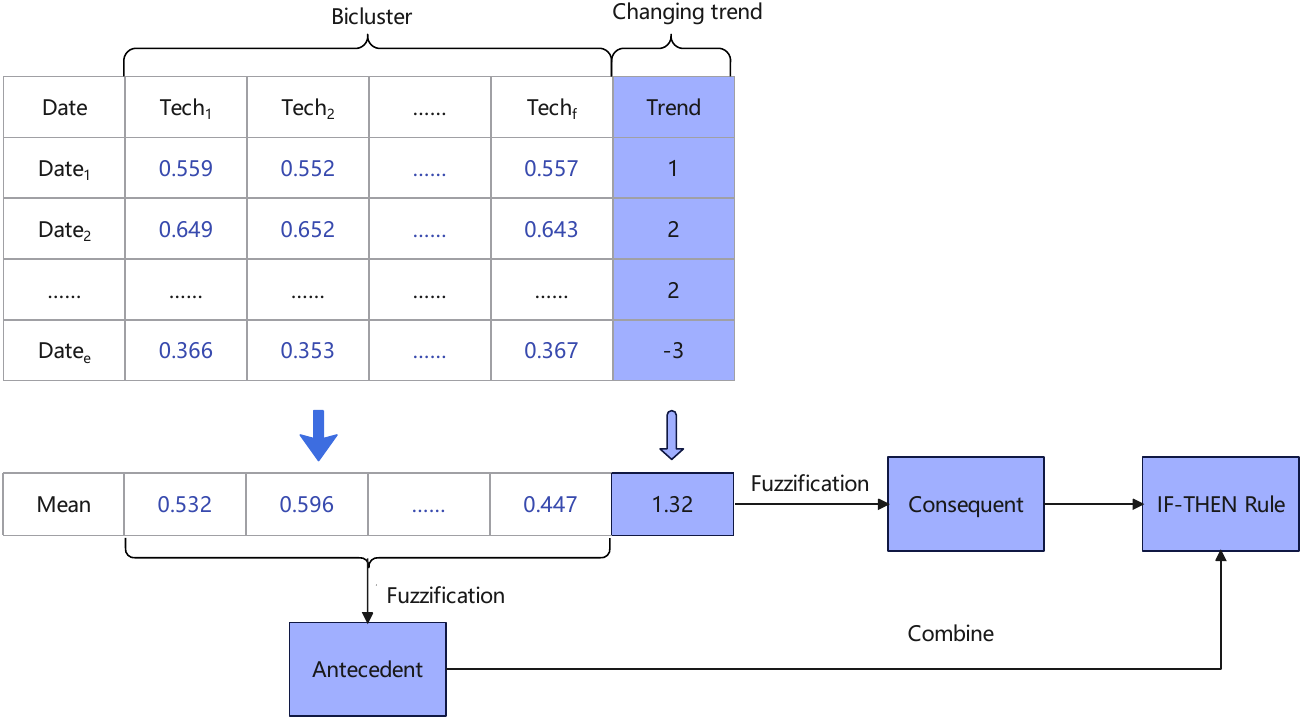}
	\caption{For each bicluster, we use the average value of each column as an input variable for the fuzzy inference system,  applying fuzzification to construct the antecedent. The trend level corresponding to each bicluster is also fuzzified to build the consequent. We combine the antecedent and the consequent into an IF-THEN fuzzy rule. Technical trading patterns are composed of biclusters and their corresponding trend levels. Each technical trading pattern can generate an IF-THEN fuzzy rule.}
	\label{tu2}
\end{figure}

The fuzzy inference system generates fuzzy rules from the technical trading patterns discovered by the BCBOF algorithm. Each pattern generates a single fuzzy rule. The modeling process of the fuzzy inference system is similar to that described in the reference \cite{huang_automated_2020}. Figure \ref{tu2} depicts the process of deriving the corresponding fuzzy rule through the fuzzification of a technical trading pattern. Within this pattern, the bicluster is represented by an $e \times f$ submatrix, where $e$ represents a subset of the trading day set $\left \{1, 2,\cdots, n  \right \} $, and $f$ represents a subset of the technical indicator set $\left \{1, 2,\cdots, p  \right \} $. The mean value of each column of this submatrix is fuzzified and utilized to construct the antecedent of the fuzzy rule. From the $n$ trading days, we select the $e$ trading days corresponding to the submatrix. We calculate the average trend level value ($Trend_i$) across these $e$ trading days. We then apply a fuzzification process to this value and utilize the resulting fuzzy set to construct the consequent of the fuzzy rule. The IF-THEN fuzzy rule is formulated as follows:
\begin{equation}  
	R^{(l)}: IF\, x_{1}\, \text{is}\, A_{1} ^{l} \; \text{and}\; x_{2}\, \text{is}\, A_{2} ^{l} \; \text{and}\; \cdots\; \text{and}\; x_{n}\, \text{is}\, A_{n} ^{l}   
\end{equation}  
\begin{equation}  
	THEN\, y\, \text{is}\, B^{(l)} 
\end{equation}
where, $x = \left(x_{1}, x_{2},\cdots, x_{f}\right)^{T} $ and $y$ represent the input and output variables of the fuzzy inference system, which are the average values of the $f$ technical indicators and the trend level. $A_{i}^{l}$ and $B^{(l)}$ are fuzzy sets. Let $M$ be the number of rules in the fuzzy rule base, i.e., $l =1,2\cdots\ M$. After obtaining the fuzzy rule base, we employ the defuzzification method \cite{huang_automated_2020} to derive a defuzzified value $\bar{y}$. 
We employ the Particle Swarm Optimization (PSO) algorithm \cite{huang_automated_2020} to obtain the trading threshold $T_h$ for each stock and identify the trading points. 

For the identified buying or selling points, we further incorporate profit-preserving and stop-loss rules to reasonably filter the buying or selling points. This ensures the effectiveness, profitability, and risk control of the trades. The profit-preserving and stop-loss rules are as follows:

$\bullet$
In financial market trading, we introduce a time delay strategy to minimize trade costs and potential risks. This strategy includes a time lag $T_{bs}$ between buying and selling points, and a time lag $T_{bt}$ between the next buying point and the previous trade. Both $T_{bs}$ and $T_{bt}$ are hyperparameters that are calculated based on historical data. This rule is designed to avoid unnecessary costs associated with frequent trading, thereby enhancing the effectiveness of the trading strategy.

$\bullet$
We require that the buying price be lower than the average closing price ($\bar{P}$) of the preceding $n$ trading days. This principle ensures entry into the market at a relatively low price, thereby maximizing potential future profits. When executing selling operations, we mandate that the selling price exceed the buying price, in order to capitalize on market fluctuations and achieve maximum investment returns. The formula for $\bar{P}$ is as follows:
\begin{equation} 
	\bar{P}_i  = \frac{1}{n}{\textstyle \sum_{m=i-n+1}^{i}} CP_m \quad 
\end{equation}
where, $\bar{P}_i$ represents the average closing price of the preceding $n$ trading days for the $i$-th trading day, and $CP_m$ represents the closing price of the $m$-th trading day.

$\bullet$
To mitigate investment risks, we have established a loss rate threshold ($T_{loss}$). If the loss rate of the invested stock exceeds this threshold, we will execute a selling operation. This operation is designed to minimize potential losses and protect the interests of investors.

By combining stock price trend predictions with profit-preserving and stop-loss rules, we derive a stock trading strategy. This trading strategy encompasses the following three types of trading rules:
\begin{equation}  
	\text{Trading Rule 1}: IF\;\bar{y}_i \ge T_h\,\; and\,\; BP_{i}< \bar{P} _i \,\;and\,\; t_{bt}\, \ge \, T_{bt}   
\end{equation}
\begin{equation}  
	THEN\;  i\; is\; a \;  buying \;  point 
\end{equation}
\begin{equation}  
	\text{Trading Rule 2}: IF\; \bar{y}_i < T_h \;and \; SP_{i}>BP \;and \; t_{bs} \,\ge\,T_{bs}   
\end{equation}
\begin{equation}  
	THEN\,  i\;is\; a \;  selling \;  point 
\end{equation}
\begin{equation}  
	\text{Trading Rule 3}: IF\;\frac{BP - SP_i}{BP} \times 100\%  \, \ge T_{loss}
\end{equation}
\begin{equation}  
	THEN\,  i\; is\; a \;  selling \;  point 
\end{equation} 
where, $\bar{y}_i$ represents the defuzzified value for the $i$-th trading day, and $T_h$ represents the trading threshold for the stock. $BP_{i}$ and $\bar{P}_i$ denote the buying price for the $i$-th trading day and the average closing price of the preceding $n$ trading days before the $i$-th day. $t_{bt}$ is the time lag between the $i$-th trading day and the previous trade. $SP_{i}$ and $BP$ represent the selling price for the $i$-th trading day and the buying price of the current trade. $t_{bs}$ is the time lag between the buying point and the selling point of this trade.

Each trading rule within the trading strategy outlines the criteria for determining buying or selling points under different circumstances. Trading Rule 1 states that we execute a buying operation when $\bar{y} \ge T_h$, provided that the buying point price is lower than the historical average price, and the time lag since the last transaction is satisfied. This helps to enter the market at a lower price, potentially increasing future profits. Trading Rule 2 states that when $\bar{y} < T_h$, it indicates unfavorable market conditions for continuing to hold the stock. Additionally, if the current selling point price is higher than the buying point price, and the time interval since the buying point is satisfied, this facilitates the realization of profits when the stock price has risen. Trading Rule 3 states that when the invested stock reaches or exceeds a predefined loss rate threshold $T_{loss}$, we immediately execute a selling operation. This helps protect investors' capital from further risks and losses. Based on these three rules, the trading strategy utilizes market prediction signals ($\bar{y}$), historical price performance ($BP$ and $\bar{P}$), trade time intervals ($T_{bt}$ and $T_{bs}$), and risk management ($T_{loss}$) to determine when to enter and exit the market.

\subsubsection{The proposed algorithm flow}\label{3.2.3}
The proposed algorithm can be divided into two parts: the training stage and the application of the system.

The specific steps of the training phase are as follows:

Step 1: We construct a training data matrix, where each row represents a different trading day and each column corresponds to various technical indicators. To ensure data consistency, we normalize each column of the matrix.

Step 2:	We employ the BCBOF algorithm to perform biclustering on the training data to obtain technical trading patterns.

Step 3: Based on the training data, we utilize the PSO algorithm to optimize the trading threshold $T_h$. Through this step, we aim to find the appropriate threshold $T_h$ for subsequent stock price trend prediction.

Step 4: We utilize the Fuzzy Inference System to convert the aforementioned trading patterns into concrete fuzzy rules, thereby constructing a fuzzy rule base.

The specific steps of system application are as follows:

Step 1: In the new data matrix, we normalize the technical indicator values for each trading day.

Step 2: We perform fuzzification on the normalized technical indicator values to determine if they match the previously established fuzzy rules.

Step 3: If the technical indicators meet the fuzzy rules, we conduct fuzzy inference. Through the defuzzification process, we obtain a defuzzified value $\bar{y}$.

Step 4: We apply the defuzzified value $\bar{y}$ to the proposed three trading rules to determine the buying or selling point.

After the aforementioned training phase and system application, the algorithm proposed in this paper can provide investors with a trading strategy based on technical indicators, which helps to improve the accuracy and efficiency of trading.

\begin{algorithm}[]
	\renewcommand{\thealgocf}{2}
	\SetAlgoLined  
	\caption{SSOBC (A stock trading strategy based on the orthogonal biclustering algorithm)} 
	\LinesNumbered 
	\KwIn{Stock training data matrix $X$, threshold $\delta$, radius $\varepsilon$, the minimum number of sample points $minPts$, new data matrix $Y$;}
	\KwOut{Fuzzy inference system;}
	normalize the training dataset matrix $X$ in \eqref{6}\;
	
	$Bicluster$ $\leftarrow$ BCBOF($X$, $\delta$, $\varepsilon$, $minPts$)\;
	\For{bicluster in Bicluster}{
		selected trading days' $Trend_i$ in \eqref{7},\eqref{8}\;
		technical pattern $\leftarrow$ $bicluster$ + selected trading days' $Trend_i$\;
			obtain the tech indicator means and trend means $\leftarrow$ technical pattern\;
			fuzzy rule $\leftarrow$ get\_fuzzy\_rule(tech indicator means, trend means)\;
			the fuzzy inference system model $\leftarrow$ fuzzy rule\;
		
	}

	$T_h \leftarrow \text{PSO\_optimize}(X)$ \cite{huang_automated_2020}\;
	normalize the new data matrix $Y$ in \eqref{6}\;
	\For{i in range(len(Y.shape[0]))}{
		$\bar{y}$ $\leftarrow$ model\_test($i$)\;
		\If{$\bar{y} \ge T_h\,\; and\,\; BP_{i}< \bar{P} _i \,\;and\,\; t_{bt}\, \ge \, T_{bt} $}{
			buying point $\leftarrow$ Trading Rule 1\;
		}
		\If{$\bar{y} < T_h\,\; and\,\; SP_{i}> BP \,\;and\,\; t_{bs} \,\ge \, T_{bs}$}{
			selling point $\leftarrow$ Trading Rule 2\;
		}
		\If{loss rate $> T_{loss}$}{
			selling point $\leftarrow$ Trading Rule 3\;
		}
	}

\end{algorithm}

\section{Experiment}\label{sec4}
In the experimental section, we validate the proposed BCBOF algorithm and the SSOBC algorithm separately. Section \ref{bcbof} tests the performance of the BCBOF algorithm. Section \ref{ssibc} examines the application of the SSOBC algorithm in the stock market.
 
\subsection{The performance of the BCBOF algorithm}\label{bcbof}  
To validate the performance of the proposed BCBOF algorithm, we compare it with several existing methods. These methods include the CC algorithm \cite{cheng_biclustering_2000}, Spectral Biclustering \cite{kluger_spectral_2003}, Plaid Model \cite{turner_improved_2005}, and LAS algorithm \cite{shabalin_finding_nodate}. To assess the quality of the biclusters generated by the algorithms, we adopted several evaluation metrics: Mean Squared Residual (MSR) \cite{cheng_biclustering_2000}, Variance (VAR) \cite{hartigan_direct_1972}, Mean Absolute Residual (MAR) \cite{yang_-clusters_2002}, Scaled Mean Squared Residual (SMSR) \cite{mukhopadhyay_novel_2009}, Relevance Index (RI) \cite{yip_harp_2004}, Maximal Standard Area (MSA) \cite{giraldez_evolutionary_nodate}, Virtual Error (VE) \cite{divina_effective_2012}. The assessment metrics we used are categorized into two groups: variance-based measures and standardization-based measures.

\subsubsection{Variance-based measures}

Mean Squared Residual (MSR) and Variance (VAR) are key biclustering quality metrics \cite{pontes_quality_2015, padilha_experimental_2019}. MSR evaluates model fit through squared residuals (lower values = better fit), while VAR measures data point variation within biclusters (lower values = greater stability). These complementary metrics provide comprehensive quality assessment. The formula for VAR is as follows:

\begin{equation}
	\text{VAR} = \sum_{i=1}^{|I|} \sum_{j=1}^{|J|} (a_{ij} - a_{IJ})^2	
\end{equation}
where, $a_{ij}$ is an element within the bicluster, $a_{IJ}$ is the mean value of the bicluster. $|I|$ and $|J|$ are the number of rows and columns in the bicluster, respectively.

The combined use of Mean Absolute Residual (MAR) and Scaled Mean Squared Residual (SMSR) enables comprehensive bicluster evaluation \cite{pontes_quality_2015, padilha_experimental_2019}. MAR assesses model precision via absolute residuals (lower values = better fit), while SMSR extends MSR by incorporating sample size effects (lower values = improved fit). Their formulas are:

\begin{equation}
	\text{MAR} = \frac{1}{|I||J|} \sum_{i=1}^{|I|} \sum_{j=1}^{|J|} \left| a_{ij} - a_{iJ} - a_{Jj} + a_{IJ} \right|
\end{equation}
\begin{equation}
	\text{SMSR} = \frac{1}{|I||J|} \sum_{i=1}^{|I|} \sum_{j=1}^{|J|} \frac{\left( a_{iJ} a_{Jj} - a_{ij} a_{IJ} \right)^2}{a_{iJ}^2 a_{Jj}^2}
\end{equation}
where, $a_{ij}$ is an element within the bicluster. $a_{iJ}$, $a_{Ij}$, and $a_{IJ}$ represent the row mean value, column mean value, and bicluster mean value, respectively. $|I|$ and $|J|$ denote the number of rows and columns in the bicluster.

The Relevance Index (RI) measures the relevance of each column by comparing the local variance to the global variance. The formula for RI is as follows:

\begin{equation}
	\text{$RI_j$} = 1 - \frac{s^2_{I_j}}{s^2_j}
\end{equation}
where, $RI_j$ is the Relevance Index, which represents the relevance of the $j$-th column. $ s^2_{I_j}$ is the local variance of that column within the bicluster, $s^2_j$ is the global variance of that column across the entire dataset. 
Total RI sums all columns' relevance, with higher values denoting better bicluster quality. For cross-algorithm comparison, we use Mean RI to account for varying column counts.

\subsubsection{Standardization-based measures}
The assessment metrics we use, including MSR, VAR, MAR, SMSR, and RI, are all variance-based. In contrast, MSA and VE are assessment metrics based on standardization \cite{pontes_quality_2015,padilha_experimental_2019}. When we obtain a bicluster $A$, we standardize it to create the standardized bicluster $A^{\prime}$. The standardization formula is as follows:
 
\begin{equation}
	a_{ij}^{\prime}  = \frac{a_{ij} - \mu_i}{\sigma_i}
\end{equation}
where, $a_{ij}^{\prime}$ is the standardized element of the bicluster. $\mu_i$ is the mean value of the $i$-th row, $\sigma_i$ is the standard deviation of the $i$-th row.

The Maximum Standard Area (MSA) assesses the data distribution within a bicluster. It quantifies the total area defined as the maximum and minimum values of adjacent columns within the bicluster. The formula for MSA is as follows:
\begin{equation}
	\text{MSA}=\sum_{j=1}^{|J|-1}\left|\frac{\max _{j}^{A^{\prime}}-\min _{j}^{A^{\prime}}+\max _{j+1}^{A^{\prime}}-\min _{j+1}^{A^{\prime}}}{2}\right| 
\end{equation}
where, $\max _{j}^{A^{\prime}}$ is the maximum value of the $j$-th column in the bicluster  $A^{\prime}$, $\min _{j}^{A^{\prime}}$ is the minimum value of the $j$-th column. Similarly, $\max _{j+1}^{A^{\prime}}$ is the maximum value of the $j$+1-th column in the bicluster $A^{\prime}$ , $\min _{j+1}^{A^{\prime}}$ is the minimum value of the $j$+1-th column. A smaller MSA value indicates a better clustering effect, which means that the data within the cluster is more uniform and consistent.

The Virtual Error (VE) measures the difference between the bicluster data and a virtual pattern. 
The formula for calculating VE is as follows: 
\begin{equation}
	\text{VE}=\frac{1}{|I||J|} \sum_{i=1}^{|I|} \sum_{j=1}^{|J|}\left|a_{ij}^{\prime}-p_{j}^{\prime}\right|
\end{equation}
where, $a_{ij}^{\prime}$ is the element in the $i$-th row and $j$-th column of the bicluster $A^{\prime}$. $p_{j}^{\prime}$ is the standardized result of the mean vector for the $j$-th column of the bicluster $A^{\prime}$. A smaller value of VE indicates that the elements of the bicluster are closer to the virtual pattern, resulting in a better clustering effect.

\subsubsection{evaluation metric results}
It is important to note that lower values of MSR, VAR, MAR, SMSR, MSA, and VE indicate better biclustering results. The RI measures the quality of biclustering by summing the correlation metrics of the columns. A higher Mean RI value indicates greater similarity among the data in the bicluster. 
Tables \ref{Yeast}-\ref{Wine-red} present the experimental results for four datasets: Yeast \cite{horton_probabilistic_1996}, Ecoli \cite{horton_probabilistic_1996}, MFCCs \cite{colonna_incremental_2015}, and Wine-red \cite{cortez_modeling_2009}. These results allow us to compare the performance of different algorithms more intuitively.
\begin{table}[htbp]
	\centering
	\caption{Experimental results for the Yeast dataset}
	\label{Yeast}
	\scriptsize
	\setlength{\tabcolsep}{3pt}
	\begin{tabular*}{\textwidth}{@{\extracolsep{\fill}} cccccccc @{}}
		\toprule
		Algorithm  & MSR & VAR  & MAR & SMSR& Mean RI &MSA&VE \\
		\midrule
		Our Method    & \bf{0.0002} & \bf{0.0004} & \bf{0.0126} & \bf{0.0004} & \bf{0.9799} &\bf{1.41421355} &0.5547 \\
		Plaid Model   & 0.0017 & 0.0129  & 0.0322 & 0.0040& 0.4057 &1.41421356 &0.7266 \\
		CC Algorithm  & 0.0014 & 0.0019  & 0.0271 & 0.0090& 0.8665 &2.38267190 &\bf{0.1604} \\	
		LAS Algorithm & 0.0035 & 0.0058  & 0.0456 & 0.0062& 0.7285 &1.41421356 &0.8169\\
		Spectral Biclustering & 0.0042 & 0.0107  & 0.0505 & 0.0388& 0.5086& 1.41421356&0.9826  \\
		\bottomrule
	\end{tabular*}
\end{table}
\begin{table}[htbp]
	\centering
	\caption{Experimental results for the Ecoli dataset}
	\label{Ecoli}
	\scriptsize
	\setlength{\tabcolsep}{3pt}
	\begin{tabular*}{\textwidth}{@{\extracolsep{\fill}} cccccccc @{}}
		\toprule
		Algorithm  & MSR & VAR  & MAR & SMSR& Mean RI &MSA&VE \\
		\midrule
		Our Method    & 0.0045 & \bf{0.0053}  & 0.0471 & 0.0165 & \bf{0.8291}&2.03233509  &\bf{0.2408}\\
		Plaid Model   & \bf{0.0002} & 0.0101  & \bf{0.0111} & \bf{0.0006}  & 0.7396&\bf{1.41421356} &0.4875 \\
		CC Algorithm  & 0.0017 & 0.0270  & 0.0299 & 0.0143 & 0.2893 &1.41421356 &0.5801 \\
		LAS Algorithm & 0.0056 & 0.0061  & 0.0590 & 0.0100 & 0.7883 &3.62619152 &0.7002  \\
		Spectral Biclustering & 0.0059 & 0.0155  & 0.0613 & 0.0615& 0.6731 &1.41421356&0.7904 \\
		\bottomrule
	\end{tabular*}
\end{table}
\begin{table}[htbp]
	\centering
	\caption{Experimental results for the MFCCs dataset}
	\label{MFCC}
	\scriptsize
	\setlength{\tabcolsep}{3pt}
	\begin{tabular*}{\textwidth}{@{\extracolsep{\fill}} cccccccc @{}}
		\toprule
		Algorithm  & MSR & VAR  & MAR & SMSR& Mean RI &MSA&VE \\
		\midrule
		Our Method  & \bf{0.0002} & \bf{0.0002}  &  \bf{0.0109} & \bf{0.0009} & \bf{0.9726}&7.19851606& \bf{0.0691}  \\
		Plaid Model & 0.0006 & 0.0010  & 0.0225 & 0.0019 & 0.8723& \bf{4.38212697} & 0.3172  \\
		CC Algorithm& 0.0018 & 0.0019 & 0.0316 & 0.0061& 0.8571&6.50887419& 0.1412  \\
		LAS Algorithm &0.0028& 0.0032  & 0.0356 & 0.0054 & 0.6427&10.1133945& 0.2024\\
		Spectral Biclustering & 0.0027 & 0.0177  & 0.0394 & 0.0213 & 0.7175 &24.9463098& 0.3891 \\
		\bottomrule
	\end{tabular*}
\end{table}

\begin{table}[htbp]
	\centering
	\caption{Experimental results for the Wine-red dataset}
	\label{Wine-red}
	\scriptsize
	\setlength{\tabcolsep}{3pt}
	\begin{tabular*}{\textwidth}{@{\extracolsep{\fill}} cccccccc @{}}
		\toprule
		Algorithm  & MSR & VAR  & MAR & SMSR& Mean RI &MSA&VE \\
		\midrule
		Our Method  & \bf{0.0001} & \bf{0.0006} & \bf{0.0079} & 0.0098   & \bf{0.9842}&0.87157187& \bf{0.1134} \\
		Plaid Model & 0.0023 & 0.0124  & 0.0398 & 0.0085  & 0.3563&\bf{6.7749e-12}& 0.2928  \\
		CC Algorithm& 0.0014 & 0.0023  & 0.0142 & \bf{0.0012}& 0.9174  &0.85941919& 0.1770  \\		
		LAS Algorithm & 0.0073 & 0.0106  & 0.0628 & 0.0175& 0.5291&0.08867402& 0.1729   \\
		Spectral Biclustering & 0.0053 & 0.0225  & 0.0526 & 0.1549& 0.3251  &2.99063824& 0.1631 \\
		\bottomrule
	\end{tabular*}
\end{table}

From Tables \ref{Yeast}-\ref{Wine-red}, we observe that the BCBOF algorithm demonstrates superior performance on the Yeast, MFCCs, and Wine-red datasets. It has significant advantages in the MSR, VAR, MAR, SMSR, and RI evaluation metrics. These metrics indicate that the BCBOF algorithm possesses strong capabilities in data fitting, maintaining stability and consistency, improving clustering accuracy, and enhancing data similarity.

Our method performs exceptionally well on the Ecoli dataset in VAR, Mean RI, and VE metrics. It indicates that we have advantages in reducing data variability, improving correlation, and lowering the error rate. However, compared to the Plaid Model, there is still room for improvement in MSR, MAR, SMSR, and MSA. Specifically, we need to enhance data fitting and maintain data consistency.
\subsubsection{other evaluation metric results}
To further validate the performance of different algorithms, Table \ref{44} compares four key metrics across methods: computational efficiency (Runtime), bicluster quantity (Biclusters), average pattern size (Avg.Size), and data coverage (Overlap). The results highlight critical trade-offs between these dimensions.
\begin{table}[htbp]
	\centering
	\caption{Experimental results for the Yeast dataset}
	\label{44}
	\scriptsize
	\setlength{\tabcolsep}{3pt}
	\begin{tabular*}{\textwidth}{@{\extracolsep{\fill}} ccccc @{}}
		\toprule
		Algorithm  & Runtime(s) & Biclusters & Avg.Size & Overlap \\
		\midrule
		Our Method  & 0.05 & 3 & 398*2 & 0.20 \\
		Plaid Model & 0.33 & 4 & 372*2 & 0.13 \\
		CC Algorithm& 0.01 & 1 & 1215*4 & 0.00 \\        
		LAS Algorithm & 6.84 & 5 & 32*2 & 0.00 \\
		Spectral Biclustering & 0.78 & 4 & 220*3 & 0.33 \\
		\bottomrule
	\end{tabular*}
\end{table}

As shown in Table \ref{44}, our method outperforms others on the Yeast dataset, achieving the best balance between speed (0.05s) and quality (3 biclusters, avg. size 398×2, overlap 0.20). Though CC runs slightly faster (0.01s) and LAS finds more biclusters (5), our approach is significantly faster than LAS (6.84s) and yields more interpretable results than CC (single zero-overlap bicluster). The moderate overlap demonstrates effective pattern detection without redundancy, making it ideal for biological data analysis where both efficiency and interpretability matter.

\subsection{Application of the SSOBC algorithm in stock trading}\label{ssibc}

In this section, we evaluate the SSOBC algorithm by comparing it with four benchmark methods: the BM-FM algorithm \cite{huang_automated_2020}, the DQN algorithm with turning point optimization \cite{wang_stock_2023}, the BIC-KNN algorithm \cite{huang_biclustering_2015}, and the buy-and-hold strategy (BHA) \cite{huang_biclustering_2015}. The DQN algorithm is a deep reinforcement learning-based trading strategy that combines LightGBM for turning point classification with Deep Q-Network (DQN) modeling. The BIC-KNN algorithm integrates biclustering with K-nearest neighbors to identify technical trading patterns, while the BHA serves as the baseline long-term holding strategy. This comparative study validates the adaptability of SSOBC in complex market environments.

Stock returns measure investment performance during holding periods, serving as a key metric for decision-making and trend analysis. We evaluate our algorithm's performance using the following return formula:
\begin{equation}
	\text{profit} = \sum_{k=1}^{M} \frac{SP_{k}-BP_{k}}{BP_{k}} \times 100\%
\end{equation}
where, $M$ is the number of trades during the test period, $SP_{k}$ is the selling price, and $BP_{k}$ is the buying price.

Table \ref{Stock Training and Testing Time} includes the 10 Shanghai and Shenzhen A-share stocks used in this article to validate the algorithm's performance, along with their training and testing times. The 10 selected A-share stocks from Shanghai and Shenzhen markets were chosen based on three criteria:
(1) Industry diversity, covering IT (SZ000034), real estate (SZ000036), consumer goods (SZ000651), and finance (SH601318);
(2) Market-cap balance, including mega-, large-, and small-to-mid-cap stocks;
(3) Cycle coverage, with training data (2020–2023) spanning pandemic shocks, policy easing, and rate hikes to test model robustness across market conditions. Regarding the selection of technical indicators, we select the same technical indicators as the reference \cite{huang_automated_2020}. The names of the indicators and their selected time periods are shown in Table \ref{Table of Technical}.

\begin{table}[ht]
	\centering
	\caption{Training data and testing data}
	\footnotesize
	\begin{tabular*}{\textwidth}{@{\extracolsep{\fill}} ccc @{}}
		
		\toprule
		\textbf{Stock code} & \textbf{Training time} & \textbf{Testing time} \\
		\midrule
		SZ000034 & 2022/07/01 - 2023/07/01 & 2024/03/19 - 2024/06/10 \\
		SZ000036 & 2022/07/01 - 2023/07/01 & 2023/07/25 - 2023/12/30 \\
		SZ000651 & 2020/03/26 - 2021/03/26 & 2021/03/30 - 2021/09/26 \\
		SZ000790 & 2020/03/26 - 2021/03/26 & 2021/04/08 - 2021/09/30 \\
		SZ000801 & 2021/05/01 - 2022/05/02 & 2022/05/12 - 2022/12/30 \\
		SH600007 & 2020/03/26 - 2021/03/26 & 2021/04/02 - 2021/09/07 \\
		SH600011 & 2022/02/01 - 2022/12/30 & 2023/01/17 - 2023/06/15 \\
		SH600017 & 2020/03/26 - 2021/03/26 & 2021/03/29 - 2021/09/26 \\
		SH600020 & 2020/03/26 - 2021/03/26 & 2021/04/02 - 2021/09/30 \\
		SH601318 & 2022/03/01 - 2023/03/01 & 2023/06/09 - 2023/10/26 \\
		\bottomrule
	\end{tabular*}
	\label{Stock Training and Testing Time}
\end{table}

\begin{table} 
	\centering
	\caption{Technical Indicators and Their Time Periods}
	\footnotesize
	\begin{tabular*}{\textwidth}{@{\extracolsep{\fill}} cccc @{}}
		
		\toprule
		\textbf{Technical indicator} & \textbf{Time period}&\textbf{Technical indicator} & \textbf{Time period} \\
		\midrule
		Williams \%R & $n=6,14,20$ & AR, BR & $n=26$\\
		ROC & $n=6,12,24,28$ & KDJ & $n=9$\\
		CCI & $n=6,12,14,28$ & SMA & $n=6,10,12,24,30$\\
		EMV & $n=6,12,14,28$ & RSI & $n=6,12,18,24,30$\\
		UOS & $n=7,14,28$ & MTM & $n=6$\\
		
		\bottomrule
	\end{tabular*}
	\label{Table of Technical}
\end{table}

\begin{figure}[h]
	\centering
	\includegraphics[height = 5cm,width = 11cm]{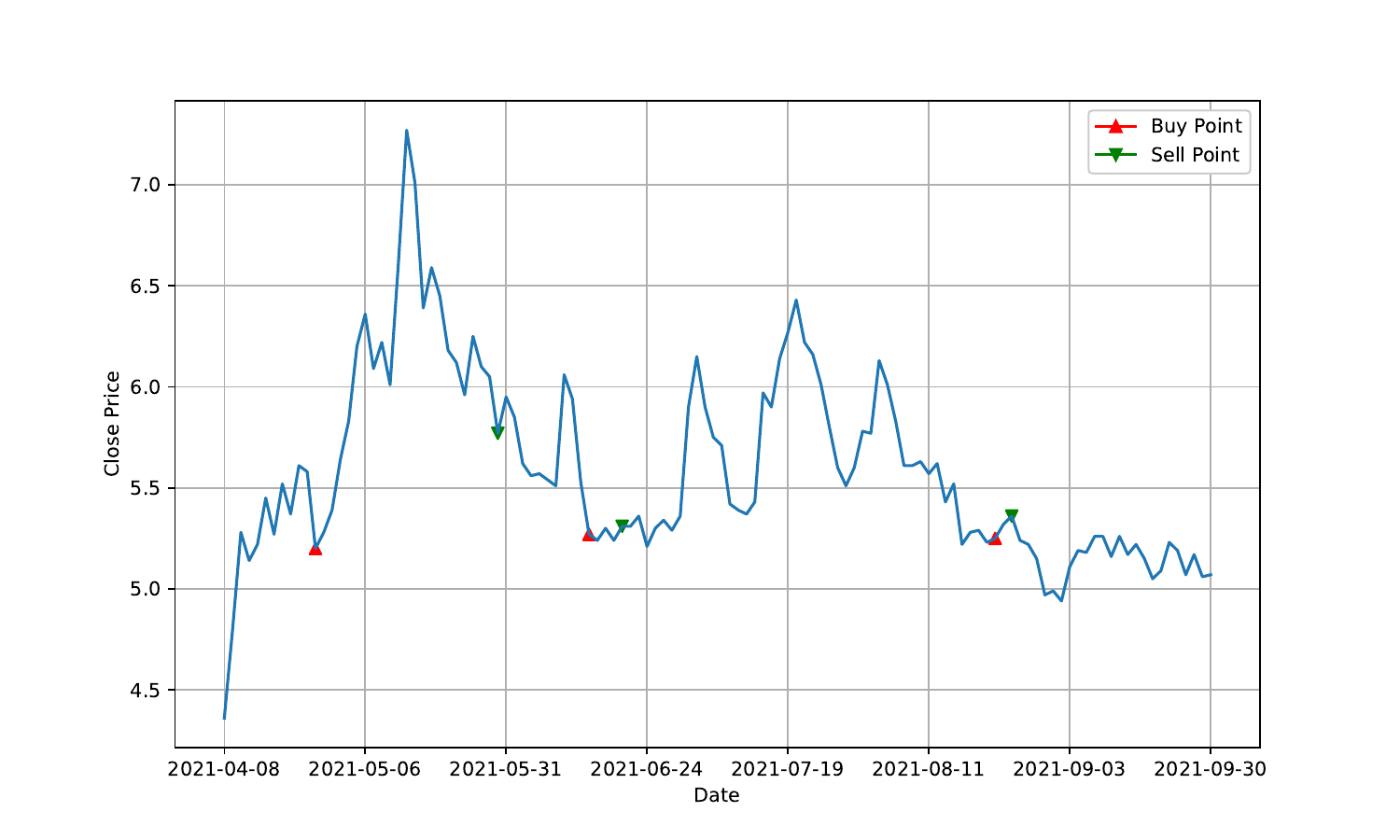}
	\caption{The BM-FM algorithm predicted three trades for the SZ000790 stock. The buying points primarily occurred within a downward channel that was characterized by lower prices. while the selling price was higher than the buying price, it was not notably high relative to the overall trend.}
	\label{2.1}
\end{figure}

\begin{figure}[h]
	\centering
	\includegraphics[height = 5cm,width = 11cm]{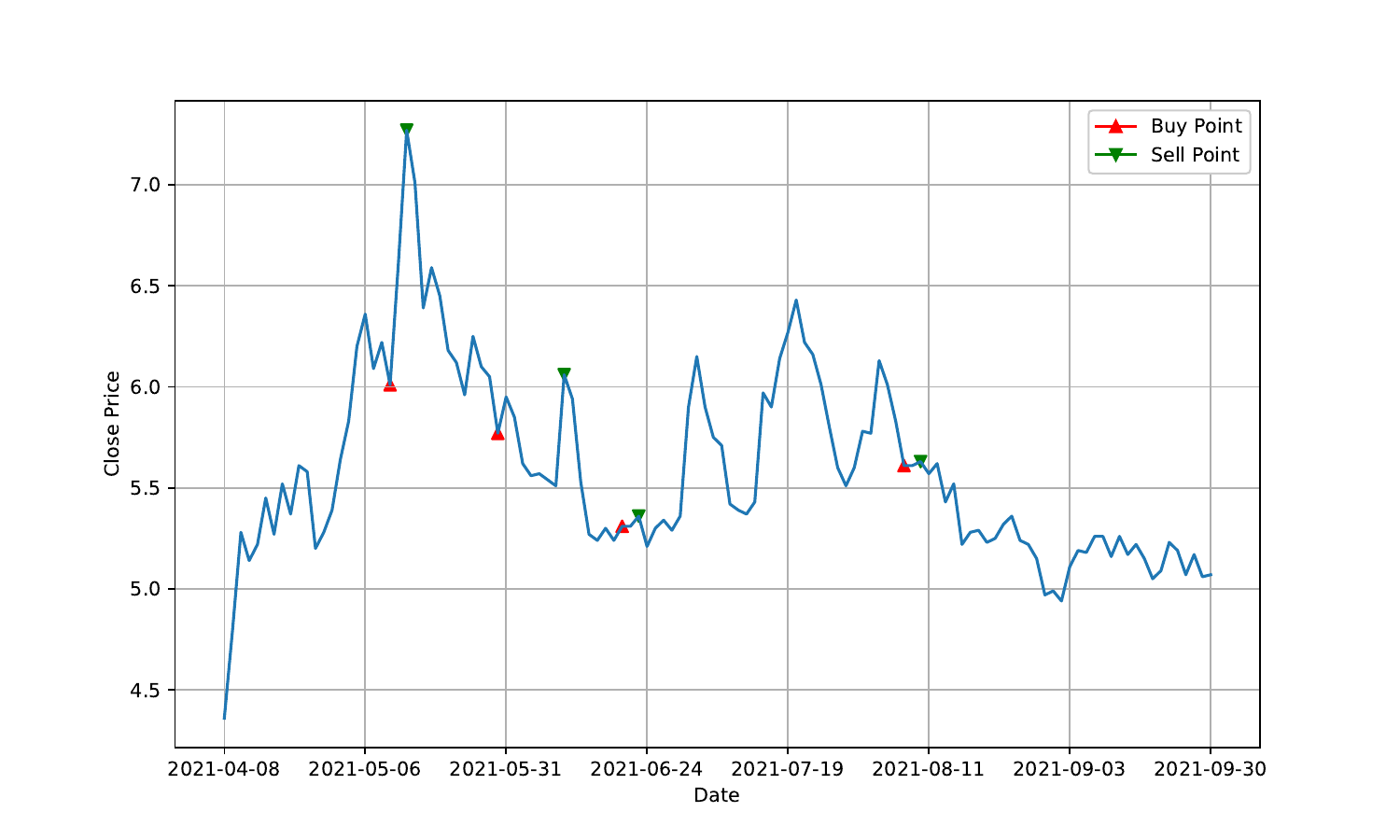} 
	\caption{The SSOBC algorithm predicted four trades for the SZ000790 stock. The price differences between the buying and selling points were large. Adhering to Trading Rule 1, which stipulated that the buying point price must be lower than the historical average, the algorithm acquired the buying points in the downward channel.}
	\label{2.2}
\end{figure}

\begin{figure}[h]
	\centering
	\includegraphics[height = 5cm,width = 11cm]{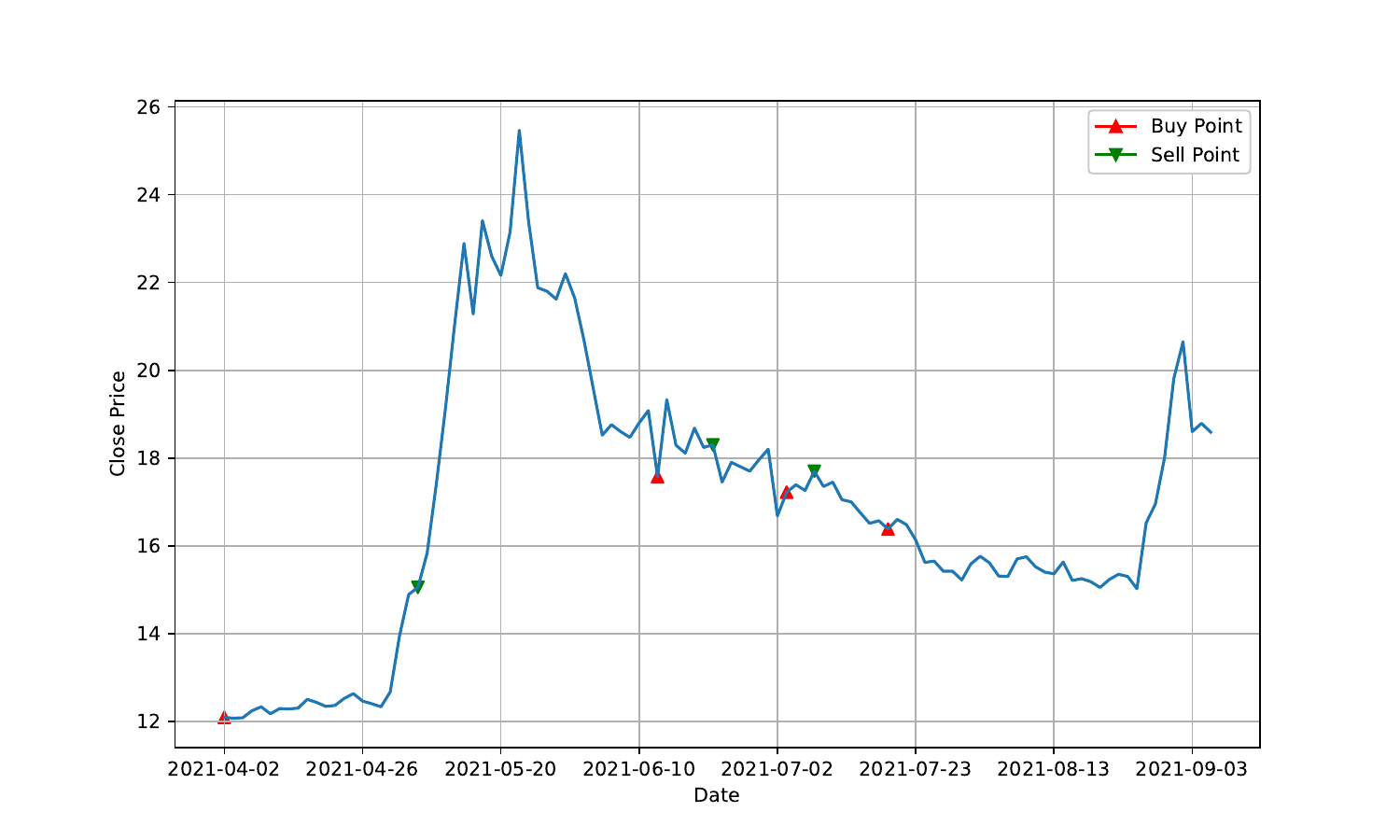}
	\caption{The BM-FM algorithm predicted three trades for the SH600007 stock. The buying point was basically the lowest point within a given period of time. However, the price differences between the buying point and the selling point were very small.}
	\label{4.1}
\end{figure}

\begin{figure}[h]
	\centering
	\includegraphics[height = 5cm,width = 11cm]{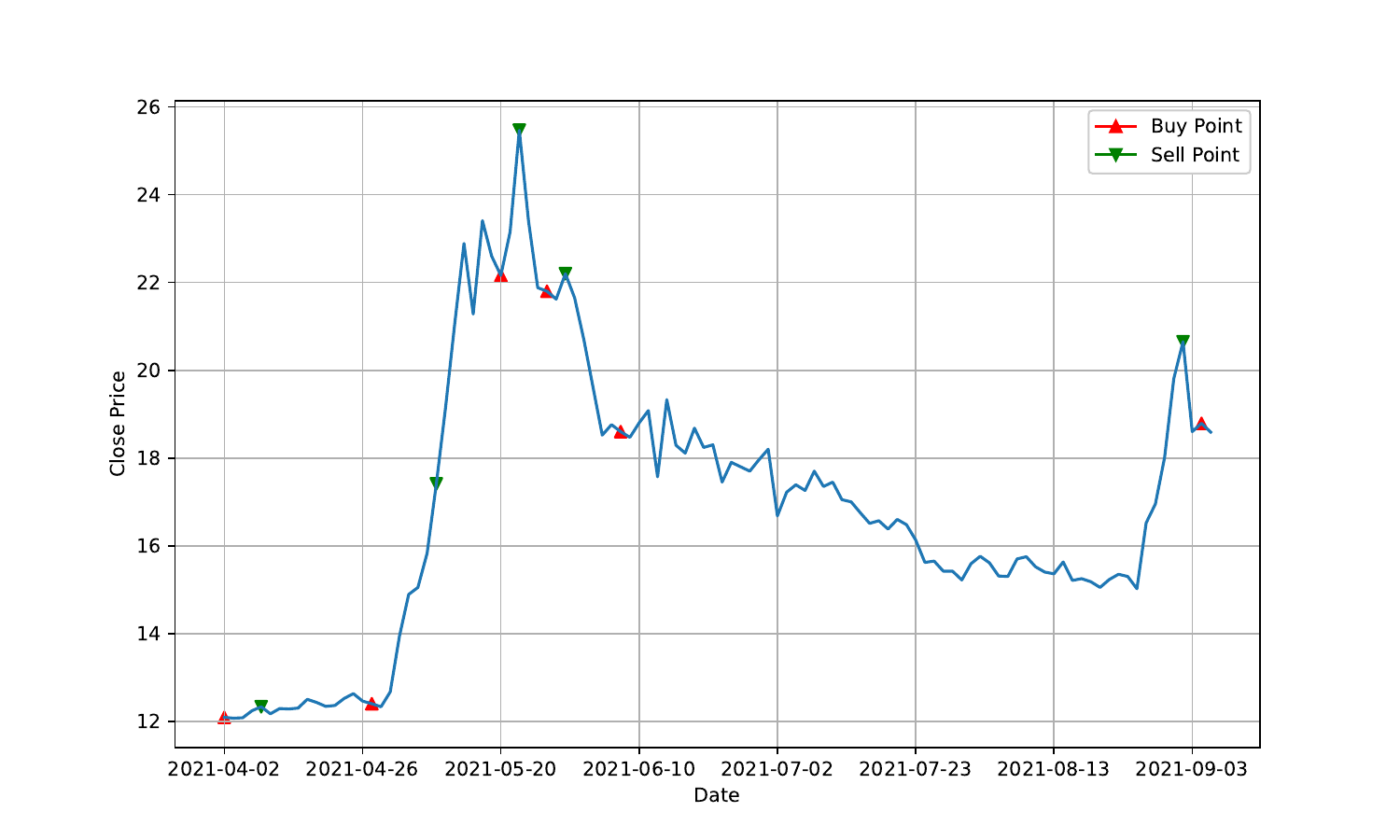}
	\caption{The SSOBC algorithm predicted five trades for the SH600007 stock. The buying and selling points were typically the troughs and peaks within a certain period of time. In the third trade, the buying point occurred within a double-top pattern, where the stock was purchased at its lowest point.}
	\label{4.2}
\end{figure}

\begin{figure}[h]
	\centering
	\includegraphics[height = 5cm,width = 11cm]{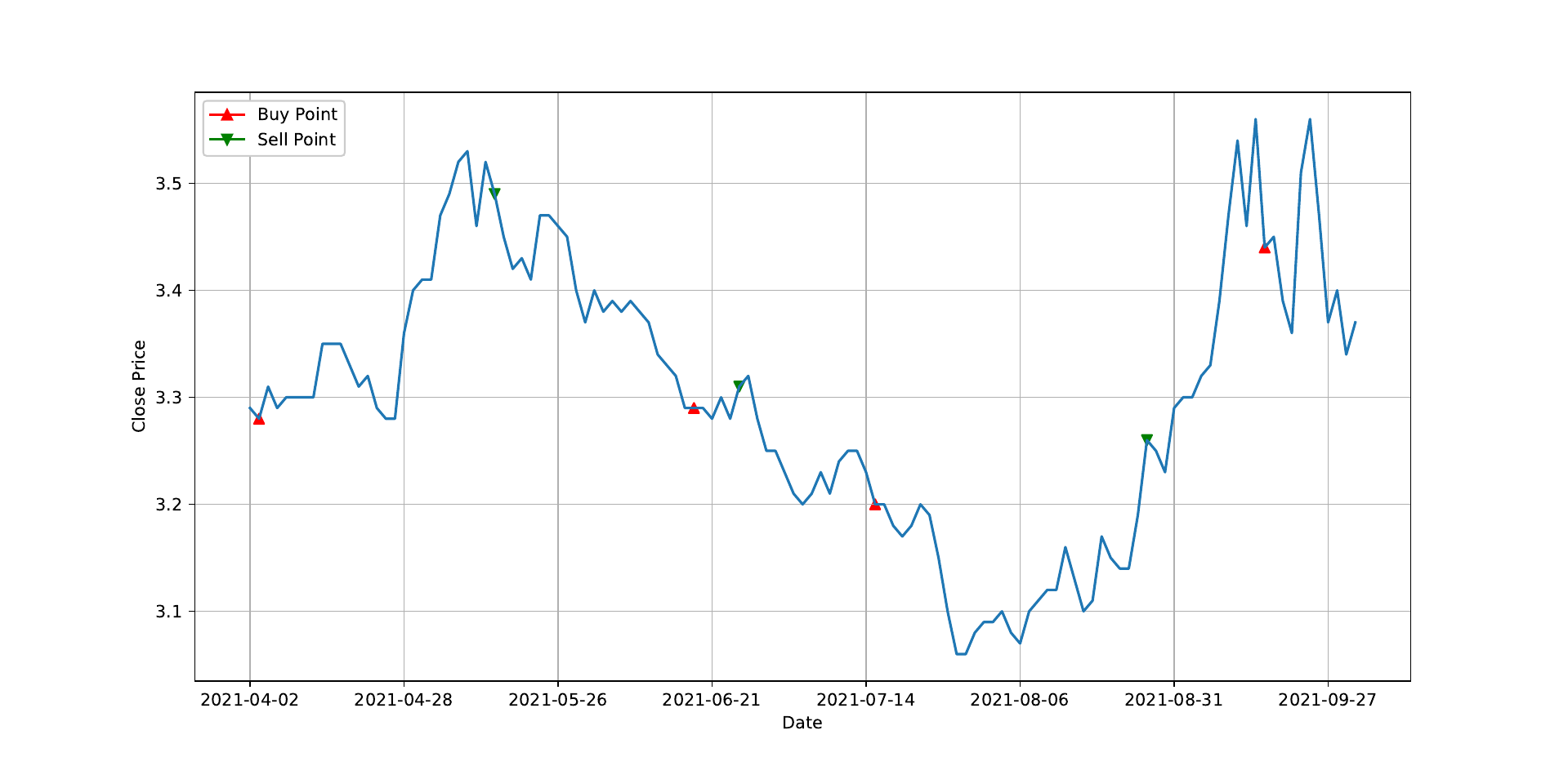}
	\caption{The BM-FM algorithm predicted three trades for the SH600020 stock. In the first trade, the stock was bought at a lower price and sold at a higher price, resulting in a better return. However, for the remaining two trades, the selling points occurred at prices similar to the buying point prices, leading to a lower return.}
	\label{9.1}
\end{figure}

\begin{figure}[h]
	\centering
	\includegraphics[height = 5cm,width = 11cm]{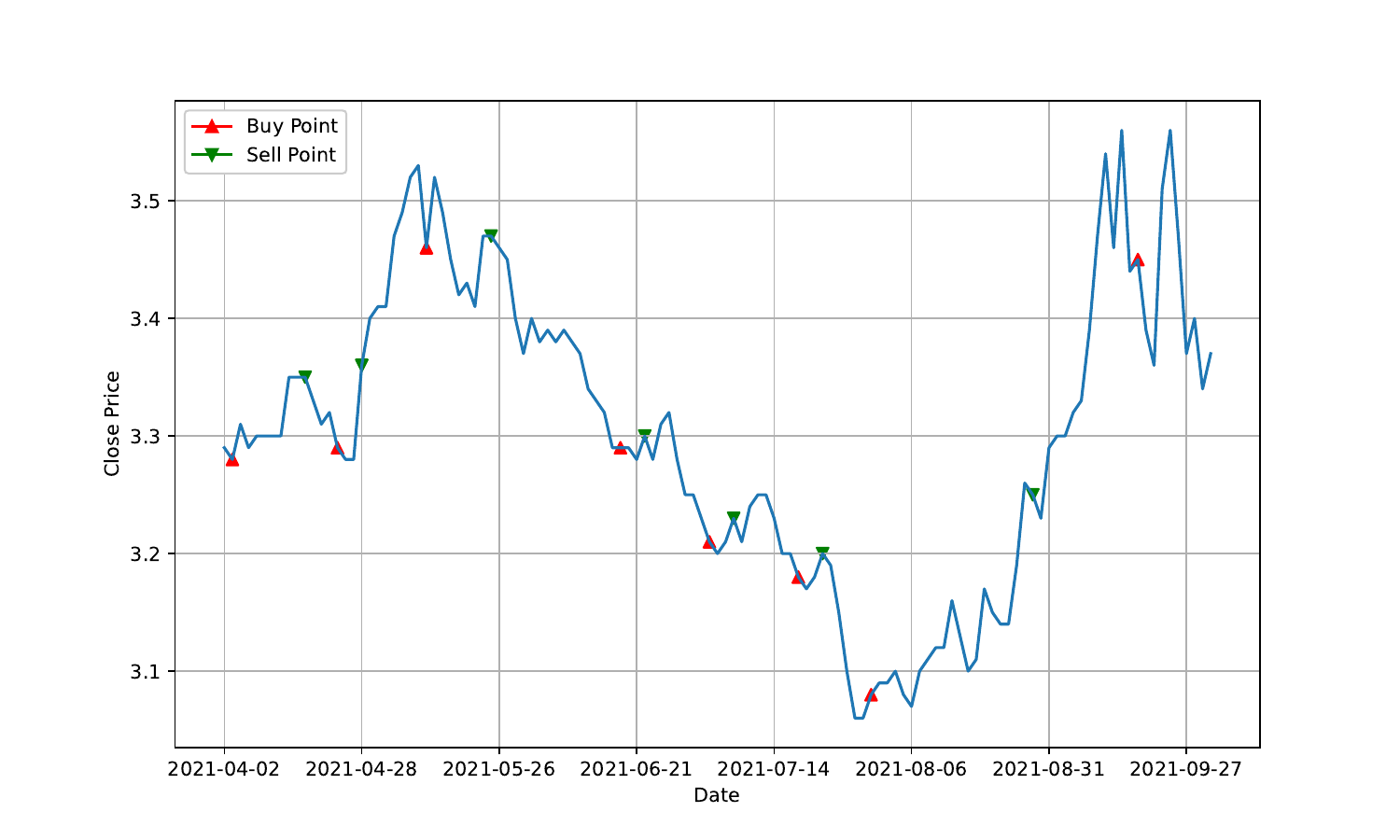}
	\caption{The SSOBC algorithm predicted seven trades for the SH600020 stock. Since Trading Rule 1 stipulated that the price of the buying point must be lower than the historical average price, the buying points were chosen within the downward channel, which could potentially improve the return. Following Trading Rule 2, the selling points were determined in the rising channel. The majority of the buying and selling points corresponded to valleys and peaks over a given period of time.}
	\label{9.2}
\end{figure}

Figure \ref{2.1} - Figure \ref{9.2} visually demonstrate the buying or selling points of the three stocks SZ000790, SH600007, and SH600020 using the BM-FM algorithm and the SSOBC algorithm. From Figure \ref{2.1} - Figure \ref{9.2}, we can find that the SSOBC algorithm has obvious advantages in identifying trade points. It not only identifies the peaks and troughs of stocks more precisely but also makes more accurate judgments in determining buying or selling points.


Taking stock SZ000790 as an example, the SSOBC algorithm effectively identifies both upward and downward trends in price fluctuations, accurately capturing price movement signals and providing trading recommendations. For SH600007, it significantly improves decision-making accuracy and reduces trading risks compared to the BM-FM method. In the case of SH600020, the algorithm successfully avoids unnecessary sell-offs during downturns while precisely identifying periodic troughs to present optimal buying opportunities for investors.

Table \ref{improved} presents the returns achieved by the SSOBC algorithm and four other algorithms across 10 stocks. The comparison results during the testing period indicate that the SSOBC algorithm outperforms the other four algorithms in terms of performance and profitability for most of the stocks. Therefore, the SSOBC algorithm exhibits potential advantages in both capturing trading opportunities and reducing risks. It also indicates the algorithm's potential to better identify effective trading signals and maintain stable performance under various market conditions. 
\begin{table}[ht]
	\centering
	\caption{Comparison of the SSOBC algorithm with other algorithms} 
	\footnotesize
	\begin{tabular*}{\textwidth}{@{\extracolsep{\fill}} cccccc @{}}
		
		\toprule
		\textbf{Stock code} & \textbf{BHA} & \textbf{BIC-KNN} & \textbf{DQN} & \textbf{BM-FM} & \parbox{2cm}{\centering \textbf{Our\\method}} \\   
		\midrule
		SZ000034 & -12.78\% & 0.31\% & 2.42\% & \bf{8.34}\% & 8.29\% \\
		SZ000036 & -8.04\% & 4.88\% & 3.17\%  & 3.95\% & \bf{7.05}\%  \\
		SZ000651 & -38.95\% & 1.01\% & 6.11\%  & 6.96\% & \bf{9.13}\%  \\
		SZ000790 & 16.28\% & 12.70\% & 25.49\%  & 13.82\% & \bf{27.29}\% \\
		SZ000801 & 1.24\% & 20.66\% &  17.87\% & 13.65\% & \bf{22.04}\%  \\
		SH600007 & 53.76\% & 41.75\% & 48.37\%  & 31.43\% & \bf{70.18}\% \\
		SH600011 & 24.50\% & 24.41\% & \bf{24.94}\% & 24.89\% & 21.15\%  \\
		SH600017 & -0.35\% & 1.74\% & 1.80\% & 1.76\% &  \bf{7.52}\% \\
		SH600020 & 2.43\% & 5.45\% & 6.94\% & 8.89\% & \bf{11.63}\%  \\
		SH601318 & -7.64\% & 5.15\% & 6.21\% & \bf{6.44}\% & 5.30\% \\
		\bottomrule
	\end{tabular*}
	\label{improved}
\end{table}

\section{CONCLUSION}\label{sec5}
  
We propose BCBOF, an enhanced biclustering algorithm that integrates traditional clustering methods with orthogonal factor models. By employing orthogonal subspace dimensionality reduction, it effectively addresses high-dimensional clustering challenges in stock data across multiple trading days and technical indicators, enabling accurate identification of technical trading patterns. These patterns are transformed into trend predictions using fuzzy rules, combined with profit-taking/stop-loss mechanisms to form a complete trading strategy. Experimental results demonstrate that BCBOF outperforms existing methods across seven biclustering evaluation metrics, while generating significantly higher investment returns than four benchmark algorithms in Shanghai and Shenzhen A-share market tests (10 stocks).

Although the proposed orthogonal factor-based biclustering algorithm improves upon traditional biclustering methods, there are still limitations that need to be addressed. Our research intentionally focuses on interpretable technical indicator patterns rather than absolute predictive accuracy. However, an important direction for the future is to develop a hybrid framework that combines the pattern recognition capabilities of our method with the feature learning advantages of deep learning, which will require the design of new heterogeneous model architectures.

\bibliographystyle{model1-num-names}
\bibliography{reference.bib}

\end{document}